\documentclass[10pt,twocolumn,letterpaper]{article}

\usepackage[pagenumbers]{cvpr} 

\usepackage[dvipsnames]{xcolor}

\usepackage{graphicx}
\usepackage{amsmath}
\usepackage{amssymb}
\usepackage{booktabs}
\usepackage{multirow}
\usepackage{color}
\usepackage{pifont}
\usepackage[T1]{fontenc}

\usepackage{dsfont}
\definecolor{cvprblue}{rgb}{0.21,0.49,0.74}
\usepackage[pagebackref,breaklinks,colorlinks,citecolor=cvprblue]{hyperref}

\def\paperID{12526} 
\def\confName{CVPR}
\def\confYear{2024}

\title{Unsupervised Template-assisted Point Cloud Shape Correspondence Network}

\author{Jiacheng Deng$^{1}$, Jiahao Lu$^1$, Tianzhu Zhang$^{1,2,}$\footnotemark[2]~ \\
\small{$^1$University of Science and Technology of China, $^2$Deep Space Exploration Lab}\\
{\tt\small \{dengjc, lujiahao \}@mail.ustc.edu.cn, {tzzhang@ustc.edu.cn}}\\
}

\begin{document}
\maketitle

\renewcommand{\thefootnote}{\fnsymbol{footnote}}
\footnotetext[2]{Corresponding Author}
\begin{abstract}
Unsupervised point cloud shape correspondence aims to establish point-wise correspondences between source and target point clouds. 
Existing methods obtain correspondences directly by computing point-wise feature similarity between point clouds.
However, non-rigid objects possess strong deformability and unusual shapes, making it a longstanding challenge to directly establish correspondences between point clouds with unconventional shapes. 
To address this challenge, we propose an unsupervised Template-Assisted point cloud shape correspondence Network, termed TANet, including a template generation module and a template assistance module. 
The proposed TANet enjoys several merits. Firstly, the template generation module establishes a set of learnable templates with explicit structures. 
Secondly, we introduce a template assistance module that extensively leverages the generated templates to establish more accurate shape correspondences from multiple perspectives. 
Extensive experiments on four human and animal datasets demonstrate that TANet achieves favorable performance against state-of-the-art methods. 
 
\end{abstract}

\section{Introduction}
Point cloud shape correspondence is a challenging task that identifies and densely matches the source and target point clouds with deformable 3D shapes. 
The task has significant implications for various industries, including augmented reality~\cite{arena2022overview}, gaming~\cite{laksono2019utilizing}, and robotics~\cite{rusu2008towards, liu2015robotic}.
However, the unrestricted mobility of humans and animals and their unusual postures have made direct correspondence between unconventional shapes a longstanding challenge in point cloud shape correspondence.

\begin{figure}[t]
    \centering
    \includegraphics[width=1\linewidth]{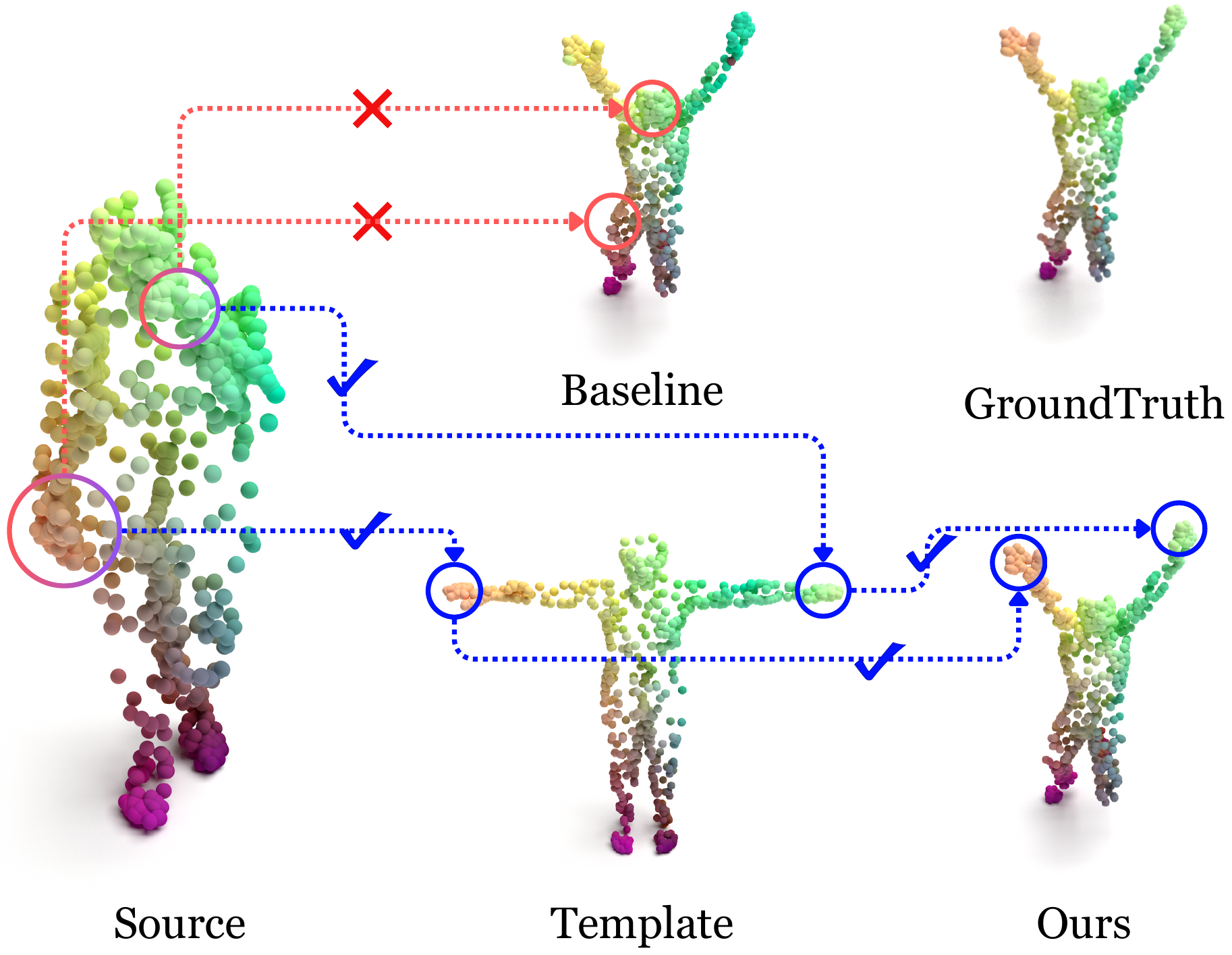}
    \vspace{-1.5em}
    \caption{
      \textbf{Visualization of template-assisted shape correspondence results. }
      Correspondences are depicted by transferring colors from the source to the target based on matching results. The baseline incorrectly aligns the hands with the head and knees between unconventional shapes. In contrast, our method leverages the template to establish accurate correspondences for the hands. }
    \label{fig:intro}
    \vspace{-1.5em}
  \end{figure}

The shape correspondence problem has been thoroughly investigated for 3D mesh data. However, such spectral-based methods~\cite{bronstein2006generalized, huang2008non, tevs2011intrinsic, ovsjanikov2012functional, litman2013learning} are mainly limited by the time-consuming pre-processing step and the reliance on mesh vertice connectivity information, frequently absent in point cloud scenarios. Point-based methods~\cite{deprelle2019learning, groueix20183d, marin2020correspondence, zeng2021corrnet3d, lang2021dpc,deng2023se,he2023hierarchical} are directly applied to the raw point cloud data, relying solely on point coordinates without connectivity information. To mitigate the resource-intensive labeling in fully supervised methods~\cite{deprelle2019learning, groueix20183d, marin2020correspondence}, the unsupervised point cloud shape correspondence task garners significant attention and yields competitive results~\cite{zeng2021corrnet3d, lang2021dpc,deng2023se,he2023hierarchical}.
DPC~\cite{lang2021dpc} designs construction losses to reduce outlier matches. HSTR~\cite{he2023hierarchical} incorporates whitening losses to mitigate the impact of shape discrepancies. SE-ORNet~\cite{deng2023se} introduces an orientation estimation module to alleviate symmetry issues.
The shapes of humans and animals are highly diverse and complex. For instance, the SURREAL dataset~\cite{groueix20183d} comprises 230K human body shapes, while the SMAL dataset~\cite{zuffi20173d} contains 10K animal shapes, including numerous unconventional forms. Establishing accurate correspondences between two unconventional point cloud shapes remains a challenging problem in shape correspondence. Current methods~\cite{zeng2021corrnet3d, lang2021dpc,deng2023se,he2023hierarchical} rely on computing the similarity between point cloud features to establish correspondences. However, in the case of unconventional point cloud pairs, the complete structure is distorted or destroyed due to limb adhesion, bending, and other postures. Without explicit structural guidance, directly calculating similarity is often insufficient to guarantee accurate correspondences. Therefore, it is essential to consider how to construct template shapes as intermediaries to assist in establishing correct correspondences. 

By analyzing prior shape correspondence methods, we have identified two pivotal issues that must be considered in shape correspondence. 
1) \textit{How to generate suitable templates for point cloud pairs?}
%
Traditional point cloud template construction~\cite{tian2020shape, yang2018foldingnet, groueix20183d} mainly relies on manual selection or autoencoders~\cite{yang2018foldingnet}. While effective for shape-stable rigid objects in the same category, balancing template complexity becomes challenging for diverse non-rigid objects of different categories. Simple templates are clear but lack structural information, risking information loss during correspondence propagation. Complex templates offer richer information but are sensitive to noise, compromising accuracy. Despite attempts~\cite{nie2023learning, zhang2021holistic} with learnable approaches, incomplete template structure remains a challenge. A more effective modeling approach is needed to generate templates suitable for intricate point cloud pairs. 
2) \textit{How to effectively leverage templates for better correspondences?}
%
Establishing direct correspondences between unconventional shapes is challenging, but building correspondences between an unconventional shape and a suitable template is relatively straightforward. A correct mapping can be formed by finding corresponding template points for the source points and identifying corresponding points on the target point cloud based on the template. Direct similarity calculations between point clouds may introduce noise and yield ambiguous similarity weights. The template allows each point in the point cloud to compute a correlation vector, providing more stable information. Seeking consensus among these correlation vectors enhances appropriate point correspondences while mitigating erroneous ones.

To achieve the above goal, we propose an \textit{Unsupervised Template-Assisted Point Cloud Shape Correspondence Network} (TANet), including a template generation module and a template assistance module. 
In the template generation module, we incorporate a template bank with several learnable templates, effectively balancing template complexity in a data-driven manner. The space aligner enriches templates with comprehensive shape structures by establishing a mapping between the template feature space and the coordinate space. The space aligner is supervised by the coordinates and encoded features of the point cloud pairs. 
In the template assistance module, the adaptive selector chooses the most suitable template from the bank based on geometric and semantic attributes. The correlation fusion process enhances point features with the template correlation vector via attention, suppressing noise and ambiguity. Moreover, we devise a transitive consistency loss to ensure coherence between the similarity computed through the template and directly computed between the source and target point clouds.

To sum up, the contributions of this work can be summarized as follows: 
(i) We propose an unsupervised template-assisted point cloud shape correspondence network, achieved by jointly designing a template generation module and a template assistance module. 
(ii) We introduce a template generation module comprising a template bank with learnable templates and the space aligner for building explicit structures. The template assistance module adaptively selects a suitable template for robust point representations and maintains transitive consistency in shape correspondences.
(iii) Extensive experiments on TOSCA~\cite{bronstein2008numerical} and SHREC'19~\cite{melzi2019shrec} benchmarks show the proposed TANet outperforming state-of-the-art methods. Cross-dataset experiments on SMAL~\cite{zuffi20173d} and SURREAL~\cite{groueix20183d} demonstrate our method's desirable generalization capabilities. 
%

\section{Related Work}
\begin{figure*}[!t]
    \vspace{-0.5em}
    \begin{center}
        \includegraphics[width=0.95\textwidth]{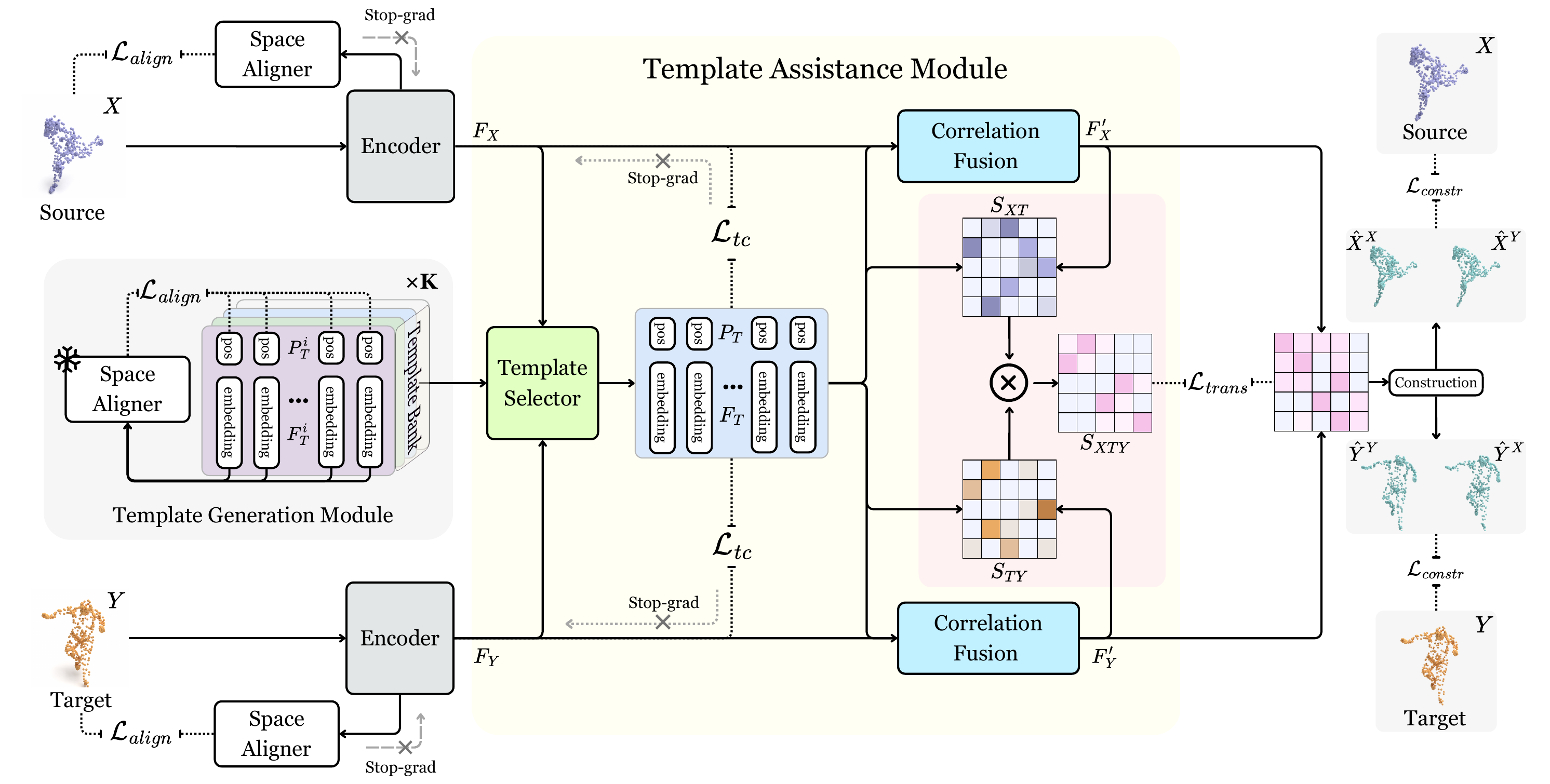}
        \vspace{-1em}
        \caption{\textbf{Illustration of the TANet. }
        TANet comprises an encoder, a template generation module, and a template assistence module. The template generation module produces several learnable shape templates in the template bank with a space aligner. The template assistance module selects suitable templates for point cloud pairs and improve accuracy via correlation fusion and transitive consistency. 
        }
        \label{pipeline}
    \end{center}
    \vspace{-2em}
\end{figure*}

In this Section, we give a brief overview of related works on point cloud shape correspondence, including deep learning for point clouds, shape correspondence, and point cloud template.

\noindent\textbf{Deep Learning for Point Clouds.}
PointNet~\cite{qi2017pointnet} employs a symmetrical function to aggregate all point features, resulting in a permutation-invariant global feature representation. 
Building upon this, PointNet++~\cite{qi2017pointnet++} takes the concept further by incorporating local information, enhancing the representation capabilities and efficiency. 
DGCNN~\cite{wang2019dynamic} introduces EdgeConv~\cite{wang2019dynamic} blocks, dynamically updating neighborhood information based on dynamic graphs, ultimately delivering improved performance in point cloud analysis. 
Similarly, KPConv~\cite{thomas2019kpconv} introduces point-wise convolution operators for point cloud feature learning. 
In the era of transformers, Point Cloud Transformer~\cite{guo2021pct} stands out as a purely global Transformer network, pioneering self-attention layers instead of encoder layers within the PointNet framework. 
Point Transformer~\cite{zhao2021point} operates within the local neighborhood of the target point cloud, enabling hierarchical extraction of geometric and semantic features.

\noindent\textbf{Shape Correspondence.}
The goal of shape correspondence is to accurately find the point-to-point correspondence between two point clouds. 
Traditional spectral-based methods~\cite{bronstein2006generalized, huang2008non, tevs2011intrinsic, ovsjanikov2012functional} utilize mesh data to project features onto Laplace-Beltrami Operator (LBO) eigenbasis and then learn a transformation between the eigendecomposition of source and target shapes, which will be translated to a dense shape correspondence. 
Cao et al.~\cite{cao2023self} transfer rich structure and connectivity information in mesh data to point cloud shape for robust matching. 
However, these methods are constrained by complex preprocessing and a reliance on vertex connectivity information. 
In contrast, point-based methods~\cite{litany2017deep, donati2020deep, corman2014supervised, sundararaman2022implicit, zeng2021corrnet3d, lang2021dpc, he2023hierarchical, deng2023se} process point clouds directly without depending on connectivity information. 
%
%
CorrNet3D~\cite{zeng2021corrnet3d} and DPC~\cite{lang2021dpc} harness the established DGCNN~\cite{wang2019dynamic} network to extract point representations through shape reconstruction. 
HSTR~\cite{he2023hierarchical} has introduced a multi-receptive-field transformer point cloud encoder that considers local point cloud structures and long-range contextual information. 
SE-ORNet~\cite{deng2023se} incorporates an orientation estimation module to align the orientations of point cloud pairs, achieving precise matching results, particularly for symmetrical parts. 
These methods directly calculate point-wise similarities, making them susceptible to matching errors caused by noise and incomplete representations. In contrast, our approach constructs and utilizes templates to establish stable correspondences. 

\noindent\textbf{Point Cloud Template.}
Point cloud templates, such as shape templates, scene priors, and semantic priors, are pivotal in various point cloud tasks, including 6D pose estimation~\cite{tian2020shape, lin2022category}, reconstruction~\cite{gkioxari2022learning, liu2022towards}, scene synthesis~\cite{nie2023learning}, and segmentation~\cite{schult2023mask3d, lu2023query, wang2022semaffinet}. 
Tian et al.~\cite{tian2020shape} develop an autoencoder trained on a collection of object models, which decodes the mean latent embedding to create shape templates. 
For similar rigid objects, generative methods can yield robust and stable templates. 
ScenePrior~\cite{nie2023learning} encodes scene priors into a latent space and utilizes sampling to synthesize plausible 3D scenes. 
SemAffiNet~\cite{wang2022semaffinet} and Mask3D~\cite{schult2023mask3d} integrate a learnable implicit semantic prior via transformer blocks with learnable queries.
However, the complex shape and category variations pose a challenge for non-rigid objects to obtain well-balanced explicit templates through generative methods. 
The learnable implicit modeling struggles to achieve explicit stuctures. 
Based on the above discussion, we propose TANet, which integrates the template generation and assistance modules within a unified framework. Through technical designs, the learnable templates not only balance shape complexity but also possess an explicitly structured shape.

\section{Method}

\subsection{Overview}
\label{Overview}
The unsupervised point cloud correspondence task strives to establish accurate one-to-one correspondences between source and target point clouds without annotations. As illustrated in Figure~\ref{pipeline}, TANet primarily comprises three key modules: a general point cloud encoder, a Template Generation Module, and a Template Assistance Module. The choice of point cloud encoder follows the previous study~\cite{he2023hierarchical}. The detailed design and optimization process for the Template Generation Module will be expounded upon in Section~\ref{Template_Bank}. Moreover, Section~\ref{template_assitance} will provide further insights into the Template Assistance Module. Finally, the training and inference process are discussed in Section~\ref{Model_Training}. 

\subsection{Template Generation Module}
\label{Template_Bank}
The template generation module consists primarily of two components: a template bank and a space aligner.

\textbf{Template bank. }
In the template bank, we integrate $K$ learnable explicit shape templates, each comprising learnable positions $P_T^i \in \mathbb{R}^{N \times 3}$ and learnable embeddings $F_T^i \in \mathbb{R}^{N \times d}$. During the training phase, the positions and embeddings of templates are updated concurrently with the network. In the inference phase, the positions and embeddings of template shapes remain unchanged, eliminating the need for additional updates based on input point clouds. The learnable embeddings $F_T^i$ parameters are initialized using Gaussian random initialization. To expedite template training, we utilize random shapes in datasets to initialize the learnable positions $P_T^i$. 
\begin{figure}[!htpb]
    \begin{center}
        \includegraphics[width=0.45\textwidth]{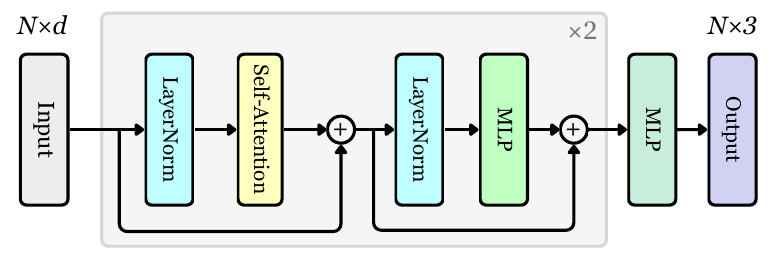}
        \vspace{-1em}
        \caption{\textbf{Space aligner structure.} The predicted point positions in the output are constrained to match the true point positions.
        }
        \vspace{-2em}
        \label{space_aligner}
    \end{center}
\end{figure}

\textbf{Space aligner. }
Space aligner $\phi$ is crafted to predict point positions through point embeddings, aiding in template learning. The specific structure is illustrated in Figure~\ref{space_aligner}. The predicted point positions undergo adjustment to align with true positions: $$\mathcal{L}_{align} = \rm{SmoothL_1} (P, \phi(F) ),$$ where $P$ are true positions, and $F$ are point embeddings. 
The space aligner is trained with the encoded feature and the coordinates from point cloud pairs.
During the inference phase, the space aligner is not involved in the computation. 

Additionally, the training of templates also includes the template construction loss $\mathcal{L}_{tc}$, which enforces consistency between the template shape and the cross-constructed template shape by point cloud pairs. Specifically, the cross-construction process is computed as follows: 
\begin{equation}
    \label{tc_construct}
    \hat{t}_{x_{i}}=\sum_{j \in \mathcal{N}_{T}\left(x_{i}\right)} \frac{e^{s_{i j}}}{\sum_{l \in \mathcal{N}_{T}(x_{i})} {e^{s_{i l}}}} t_{j},
\end{equation}
where $x_i \in X$, $t_j \in P_T$ and $s_{ij}\in S_{XT}$. $S_{XT}$ is the similarity matrix between $F_X$ and $F_T$. $\mathcal{N}_{T}(x_{i})$ represents the latent $k$-nearest neighbors of $x_{i}$ in the target $F_T$. The cross-construction of $P_T$ by the source point cloud $X$ is denoted $\hat{T}_{X} \in \mathbb{R}^{N \times 3}$, where $\hat{T}_{X}^i = \hat{t}_{x_{i}}$. 
After the introduction of the cross-construction process, the template construction loss $\mathcal{L}_{tc}$ can be computed as follows: 
\begin{equation}
    \label{tc_loss}
    \mathcal{L}_{tc} = \mathbf{CD}(P_T, \hat{T}_{X})+ \mathbf{CD}(P_T, \hat{T}_{Y}),
\end{equation}
where $\mathbf{CD}$ denotes chamfer distance. 
It is essential to highlight that the point encoder remains unaltered during the backpropagation of $\mathcal{L}_{align}$ and $\mathcal{L}_{tc}$ gradients, preventing interference with the feature extraction of point cloud pairs.

\subsection{Template Assistance Module}
\label{template_assitance}
The template assistance module aims to enhance the correspondence accuracy in point cloud pairs through templates. It comprises three primary components: Template Selector, Correlation Fusion, and Transitive Consistency. 

\begin{figure}[!htpb]
    \begin{center}
        \includegraphics[width=0.45\textwidth]{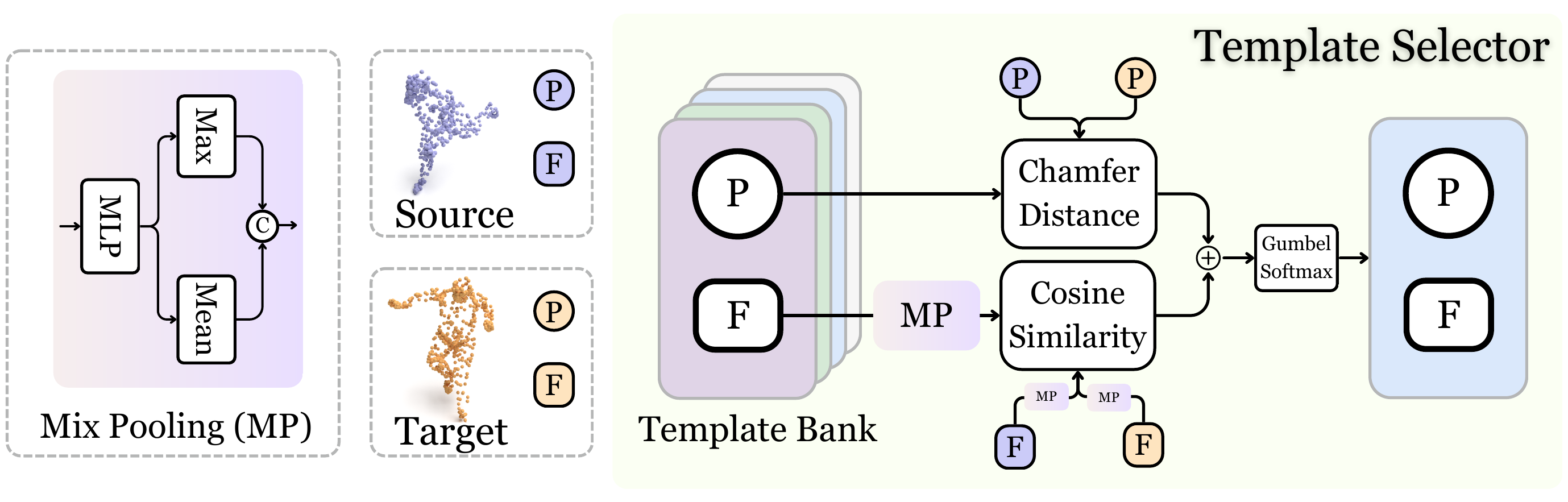}
        \vspace{-1em}
        \caption{\textbf{Template selector structure.} The best suitable template is determined through geometric and semantic measures of similarity between templates and point cloud pairs.
        }
        \vspace{-2em}
        \label{template_selector}
    \end{center}
\end{figure}
\textbf{Template selector.}
The input to the Template Selector comprises point cloud coordinates $X \& Y \in \mathbb{R}^{N \times 3}$ and features $F_X \& F_Y \in \mathbb{R}^{N \times d}$, along with template point cloud coordinates $\{P_T^i|i=1,2,\ldots, K\}$ and embeddings $\{F_T^i|i=1,2,\ldots, K\}$ from the template bank. Geometric and semantic similarities between the point cloud pair and templates guide the adaptive template selection. As illustrated in Figure~\ref{template_selector}, it involves calculating the chamfer distance as the geometric similarity between the point cloud pair and template point cloud. To get semantic similarity, the point cloud features and template embeddings undergo mix pooling and cosine similarity computation. The fusion of geometric and semantic similarities is employed in Gumbel-softmax to ascertain the most fitting template $T$ for the given point cloud pair. The specific computation formula is as follows:
{\small
\begin{equation}
    \begin{aligned}
    \label{selector}
    T &= \mathbf{GS} \{ (\mathbf{MP}(F_T^i) \cdot \mathbf{MP}(F_X)+\mathbf{MP}(F_T^i) \cdot \mathbf{MP}(F_Y)) \\
      &+ (\mathbf{CD}(P_T^i, X)+\mathbf{CD}(P_T^i, Y)) | i=1,2,\ldots, K  \},
    \end{aligned}
\end{equation}
}
where $\mathbf{GS}$ represents Gumbel-Softmax, $\mathbf{MP}$ denotes Mix Pooing, and $\mathbf{CD}$ stands for chamfer distance.

\begin{figure}[!htpb]
    \begin{center}
        \includegraphics[width=0.45\textwidth]{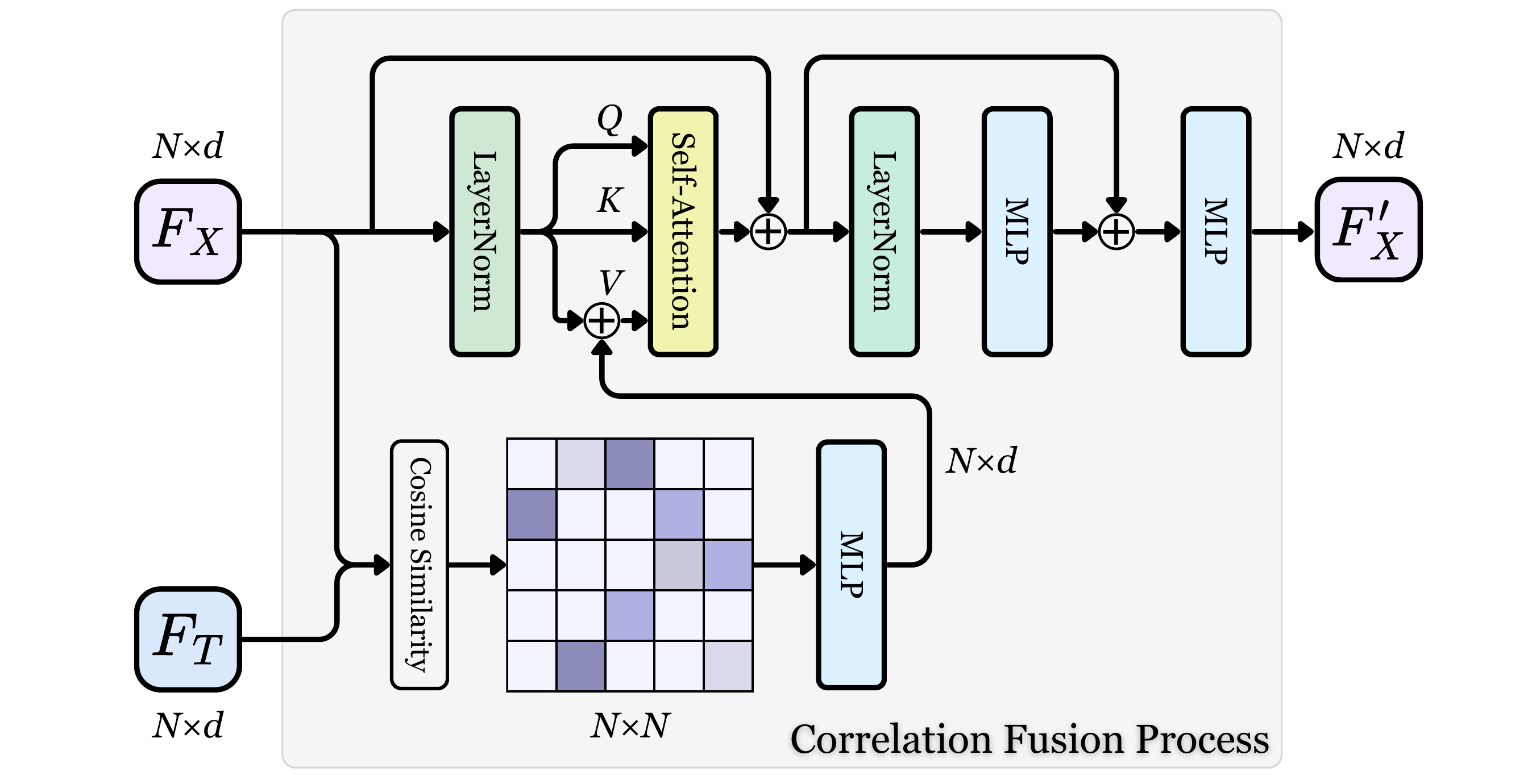}
        \vspace{-1em}
        \caption{\textbf{The correlation fusion process.} Features of point cloud pairs and embeddings of templates compute correlation vectors, which are then fused using an attention mechanism.
        }
        \vspace{-2em}
        \label{correlation_fusion}
    \end{center}
\end{figure}

\textbf{Correlation fusion.}
As depicted in Figure~\ref{correlation_fusion}, template embeddings $F_T$ and point cloud features $F_X/F_Y$ are input into the correlation fusion process to enhance point cloud features. Initially, a correlation between the point cloud and the template is established. Subsequently, an attention mechanism incorporates the correlation vectors into point cloud features, yielding $F'_X/F'_Y \in \mathbb{R}^{N \times d}$. The similarity in the attention mechanism is computed based on point cloud features to underscore the spatial distribution of shapes, enhancing perception to spatial relationships. The introduction of template correlation aids in refining point cloud features and mitigating ambiguous representations in certain noisy point clouds. 

\begin{figure}[!htpb]
    \begin{center}
        \includegraphics[width=0.38\textwidth]{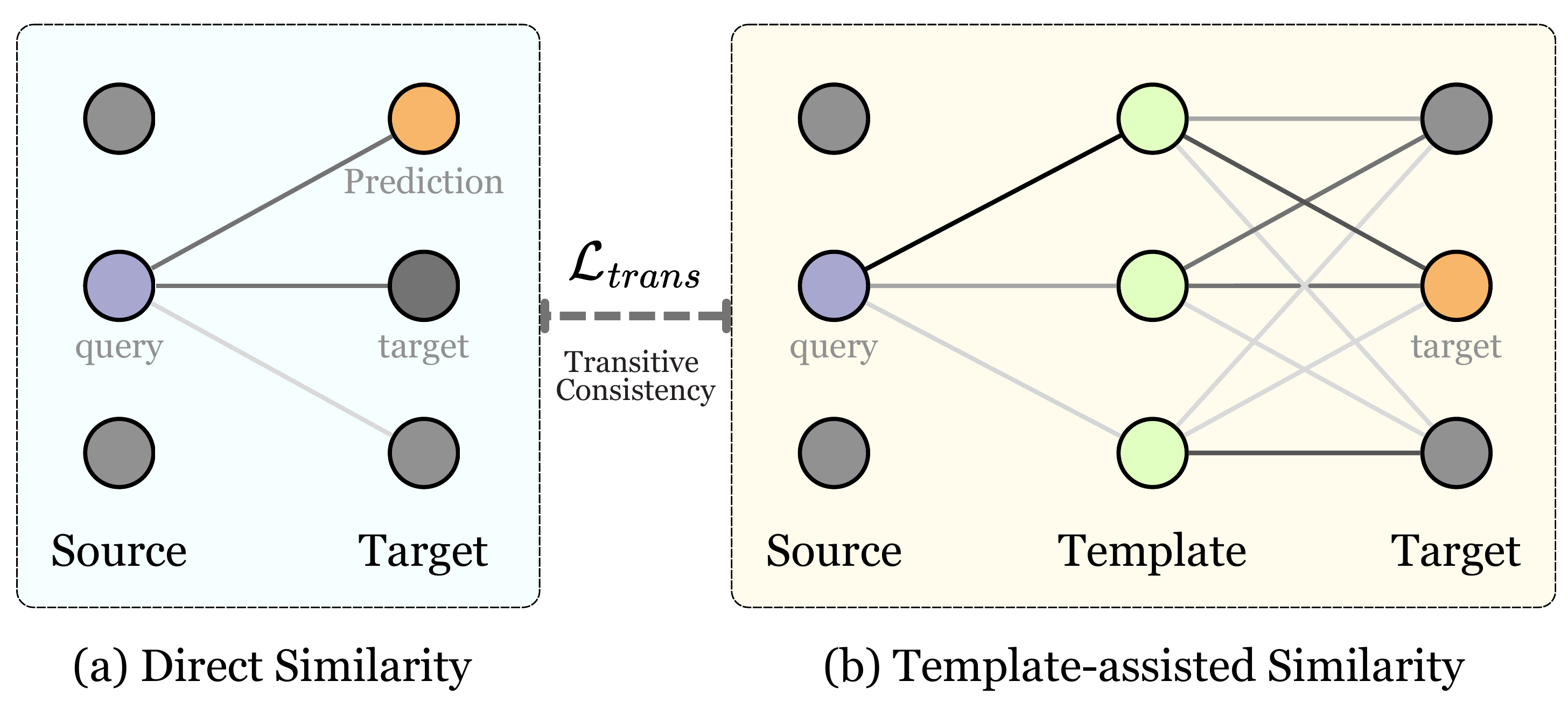}
        \vspace{-1em}
        \caption{\textbf{Transitive Consistency.} (a) the direct similarity computation between the source and target point clouds.  (b) a stable template is employed to transmit the similarity relation. $L_{trans}$ ensures the consistency between these two similarity results.
        }
        \vspace{-2em}
        \label{transitive}
    \end{center}
\end{figure}

\textbf{Transitive consistency.}
Utilizing the fused source features $F'_X$ and template embeddings $F_T$, we compute the similarity matrix $S_{XT}$, and $S_{TY}$ is obtained by calculating the similarity between $F_T$ and $F'_Y$. The multiplication of $S_{XT}$ and $S_{TY}$ matrices yields the similarity matrix $S_{XTY}$, reflecting the similarity between the source and target point clouds transmitted through the template. 
As shown in Figure~\ref{transitive}, (a) Direct similarity illustrates the process of directly calculating the similarity between the source point cloud and the target point cloud: 
\begin{equation}
    \label{direct_sim}
    S_{XY} = {F'_X} {F'_Y}^\top.
\end{equation}
In contrast, (b) Template-assisted similarity demonstrates the computation of $S_{XTY}$. The multiplication of matrices $S_{XT}$ and $S_{TY}$ essentially results in an indirect computation of the similarity relationship between the source and target point clouds. This process involves a weighted sum of the similarity calculated between the template point cloud and the target point cloud, providing a more stable and accurate similarity result for source and target points. The Template-assisted similarity method helps mitigate the impact of outliers and noise on the similarity calculation, leading to more reliable results.
In order to guide the establishment of more stable and accurate similarity relationships between point clouds without compromising the inference speed, we introduce the Transitive Consistency loss ($\mathcal{L}_{trans}$), promoting the consistency between $S_{XTY}$ and the similarity matrix $S_{XY}$. 
The Transitive Consistency loss is calculated as follows:
\begin{equation}
    \label{trans_loss}
    \mathcal{L}_{trans} = \mathbf{CE}(S_{XTY}, S_{XY}),
\end{equation}
where $\mathbf{CE}$ means Cross-Entropy loss. 

\subsection{Model Training \& Inference}
\label{Model_Training}
Following the previous work~\cite{lang2021dpc,deng2023se,he2023hierarchical}, we also incorporate the construction loss ($\mathcal{L}_{constr}$), encompassing both cross-construction and self-construction between the source shape $X$ and the target shape $Y$. The construction loss promotes feature smoothness and unique correspondences. The computation process and formula for cross-construction and self-construction are similar to Equation~\ref{tc_construct}. Specific formulas are detailed in the supplementary material. 
To sum up, the total loss of TANet is formulated as:
\begin{equation}
\begin{aligned}
    \label{overall_loss}
    \mathcal{L}_{total}= \lambda_1 \mathcal{L}_{trans} + \lambda_2 \mathcal{L}_{align}  + \lambda_3 \mathcal{L}_{tc} + \lambda_4 \mathcal{L}_{constr},
\end{aligned}
\end{equation}
where $\lambda_i$ are hyperparameters, balancing the contribution of different loss terms.
The training of both the template and the network is jointly achieved, enabling end-to-end training without requiring additional prolonged iterations. 

During inference, we set the closest point $y_{j^*}$ in the feature space for each point $x_i$ as its corresponding point.
This selection rule can be formulated as:
\begin{equation}
        \label{inference}
        f\left(x_{i}\right)=y_{j^{*}}, j^{*}=\underset{j}{\operatorname{argmax}} (s_{i j}).
        \vspace{-1em}
\end{equation}

\section{Experiments}
\subsection{Experimental Setup}
\textbf{Datasets.}
We evaluate our method on the most popular benchmarks, including the TOSCA~\cite{bronstein2008numerical} dataset and the SHREC'19~\cite{melzi2019shrec} dataset.
TOSCA comprises 41 different shapes of various animal species. Following previous methods~\cite{lang2021dpc,he2023hierarchical}, we pair these shapes to create both the training and testing sets. 
SHREC'19 consists of 44 real human models paired into 430 shape correspondence samples.
SURREAL~\cite{groueix20183d} and SMAL~\cite{zuffi20173d} are two large-scale datasets that we employ to evaluate the generalization capabilities of our method.
The SURREAL dataset comprises 230,000 training shapes, from which we randomly sample and pair 2,000 shape pairs for training. 
The SMAL dataset includes parameterized animal models for generating shape pairs to train shape correspondence models. 
Furthermore, robustness experiments are conducted on the real-scanned Owlii dataset~\cite{xu2017owlii} and the partial shape dataset SHREC'16~\cite{cosmo2016shrec}. 
Besides, we maintain a consistent point count of $n=1024$ in accordance with prior studies~\cite{zeng2021corrnet3d, lang2021dpc, he2023hierarchical}.

\textbf{Evaluation Metrics.}
The primary evaluation metrics for the point cloud shape correspondence task include two key measures: average correspondence error and correspondence accuracy. The average correspondence error measures a source and target shape pair $(X,Y)$ as follows:

\begin{equation}
\label{correspondence error}
err=\frac{1}{n} \sum_{x_{i} \in X}\left|f\left(x_{i}\right)-y_{gt}\right|_{2},
\end{equation}
where $y_{gt} \in Y$ represents the ground-truth matching point to $x_{i}$.
Correspondence accuracy is formulated as follows:
\begin{equation}
\label{correspondence accuracy}
acc(\epsilon)=\frac{1}{n} \sum_{x_{i} \in X} \mathds{1}\left(\left|f\left(x_{i}\right)-y_{gt}\right|_{2}<\epsilon d\right),
\end{equation}
where $\mathds{1} (\cdot)$ denotes the indicator function, $d$ is the maximum Euclidean distance between points in $Y$, and $\epsilon \in [0, 1]$ signifies the error tolerance.

\textbf{Implementation details.}
The number of learnable templates ($K$) in the Template Bank is set to 4, with a template embedding dimension of 512. The Space Aligner incorporates two attention blocks, each with a dimensionality of 512. We employ a transformer encoder, akin to existing method~\cite{he2023hierarchical}, featuring four layers of attention blocks as our point cloud encoder. For the Correlation Fusion process, the dimensions of self-attention and MLP are set to 512. In Equation~\ref{overall_loss}, $\lambda_{1}$, $\lambda_{2}$, $\lambda_{3}$, and $\lambda_{4}$ are set to 0.5, 0.5, 1, and 1. All experiments with our method on various datasets are conducted on a single GeForce RTX 3090 device using the PyTorch 1.10.1 framework~\cite{paszke2017automatic}. Model training utilizes the AdamW~\cite{loshchilov2017decoupled} optimizer with a learning rate of 5e-4, a weight decay of 5e-4, and a batch size of 4.

\begin{table}[!htbp]
  \begin{center}
    \footnotesize
    \setlength\tabcolsep{6pt}
    \caption{\textbf{Comparison on TOSCA and SHREC'19 benchmarks.} Acc signifies correspondence accuracy at 0.01 error tolerance, and err denotes average correspondence error. Higher accuracy and lower error reflect a better result, with the best and second-best outcomes highlighted in bold and underlined, respectively.}
    \label{table:small_dataset}
    \vspace{-0.8em}
    \begin{tabular}{c|c|cc|cc}
      \toprule
      \multirow{2}*{Method} & \multirow{2}*{Input} & \multicolumn{2}{c|}{TOSCA} & \multicolumn{2}{c}{SHREC'19} \\
      \cline{3-4}\cline{5-6}
      & & acc $\uparrow$ & err $\downarrow$ & acc $\uparrow$ & err $\downarrow$ \\
      \midrule
      {\color{gray}SURFMNet\cite{roufosse2019unsupervised}}   & {\color{gray}Mesh} &  {\color{gray}/} & {\color{gray}/} & {\color{gray}5.9\%} & {\color{gray}0.2} \\
      \midrule
      CorrNet3D\cite{zeng2021corrnet3d}         & Point & 0.3\%  & 32.7 & 0.4\% & 33.8  \\
      DPC\cite{lang2021dpc}                     & Point & 34.7\%  & 2.8 & 15.3\% & 5.6 \\
      SE-ORNet\cite{deng2023se}                 & Point & 38.3\%  & 2.7 & 17.5\% & 5.1 \\
      HSTR\cite{he2023hierarchical}             & Point & \underline{52.3\%}  & \underline{1.2} & \underline{19.3\%} & \underline{4.9} \\
      \textbf{Ours}                             & Point & \textbf{65.1\%} & \textbf{0.7} & \textbf{21.5\%} & \textbf{4.5} \\
      \bottomrule
    \end{tabular}
    \vspace{-2em}
  \end{center}
\end{table}

\subsection{Comparison with State-of-the-art Methods}
We conducted comprehensive experimental comparisons with existing methods on the TOSCA animal dataset and the SHREC'19 human dataset. Furthermore, to ensure a fair comparison with existing methods, our approach is implemented without the use of post-processing or additional connectivity information. 

\begin{figure}[!htbp]
  \begin{center}
      \includegraphics[width=0.48\textwidth]{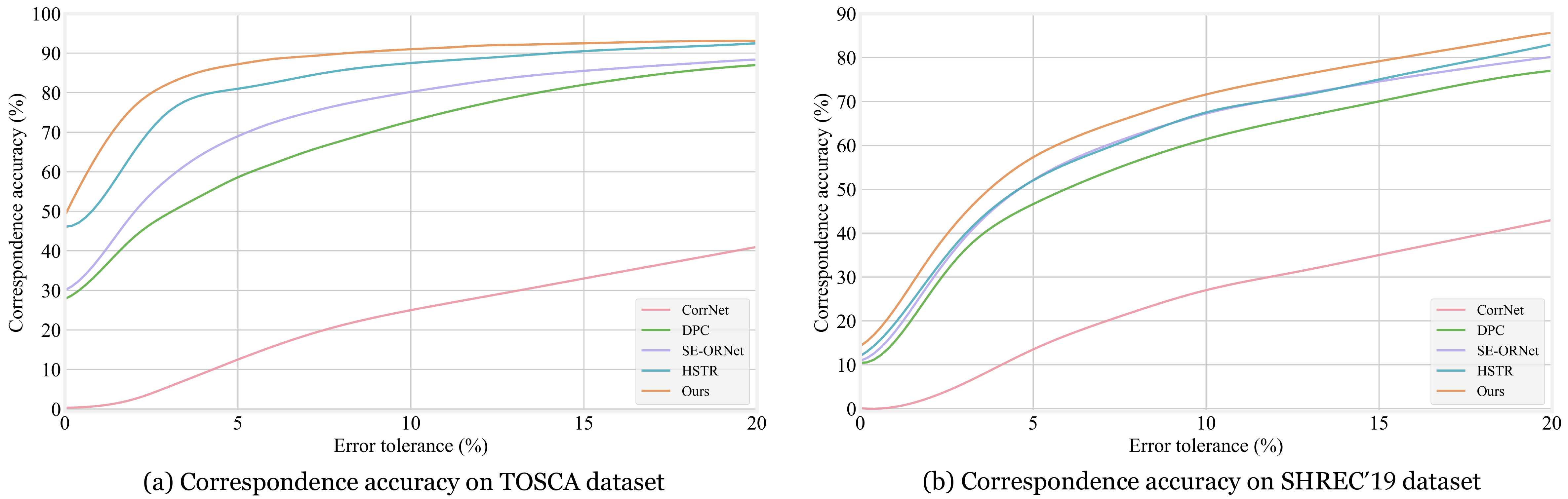}
      \vspace{-2em}
      \caption{\textbf{Correspondence accuracy at various error tolerances. } (a) Corresponding accuracy of different methods on the TOSCA dataset under various error tolerances. (b) Experiments similar to (a) conducted on the SHREC'19 dataset.}
      \vspace{-1.5em}
      \label{acc_graph}
  \end{center}
\end{figure}

\textbf{Evaluation on TOSCA dataset.}
As shown in Table~\ref{table:small_dataset}, our approach demonstrates significant improvement on the TOSCA dataset, achieving a new state-of-the-art performance with a 12.8\% increase in accuracy and a minimal average error of 0.7cm. For a comprehensive depiction of accuracy across different error tolerances, we present correspondence accuracy in Figure~\ref{acc_graph}(a). Similarly, our method outperforms others at different error tolerances.

\textbf{Evaluation on SHREC'19 dataset.}
As indicated in Table~\ref{table:small_dataset}, our method demonstrates significant improvement on the SHREC'19 dataset, achieving an impressive 2.2\% increase in accuracy and a reduction of 0.4cm in error, consequently establishing a new state-of-the-art performance. Figure~\ref{acc_graph}(b) illustrates the correspondence accuracy of various competitive methods at different error tolerances, showcasing our method's clear superiority across different levels of error tolerance.

\begin{table}[!htbp]
  \begin{center}
    \footnotesize
    \setlength\tabcolsep{5pt}
    \centering
    \caption{\textbf{Evaluation of the model with different designs on TOSCA dataset.} TGM denotes the template generation module, $L_{tc}$ stands for template construction loss, TS represents template selector, CF denotes correlation fusion process, and $L_{trans}$ is the transitive consistency loss.}
    \vspace{-0.8em}
    \begin{tabular}{c|ccccc|cc}
      \toprule
      & \multirow{2}*{TGM} &  \multirow{2}*{$\mathcal{L}_{tc}$} &   \multirow{2}*{TS} & \multirow{2}*{CF} & \multirow{2}*{$\mathcal{L}_{trans}$} & \multicolumn{2}{c}{TOSCA} \\
      \cline{7-8}
      & & & & & & acc $\uparrow$ & err $\downarrow$ \\
      \midrule
      {[A]} & \ding{55}&\ding{55}&\ding{55}&\ding{55}&\ding{55}&52.3\%&1.2\\
      {[B]} & \ding{51}&\ding{55}&\ding{55}&\ding{51}&\ding{55}&54.6\%&1.2\\
      {[C]} & \ding{51}&\ding{55}&\ding{55}&\ding{55}&\ding{51}&55.8\%&1.1\\
      {[D]} & \ding{51}&\ding{51}&\ding{55}&\ding{51}&\ding{55}&59.3\%&1.0\\
      {[E]} & \ding{51}&\ding{51}&\ding{55}&\ding{55}&\ding{51}&58.8\% &1.0\\
      {[F]} & \ding{51}&\ding{51}&\ding{51}&\ding{51}&\ding{55}&61.9\%&0.9\\
      {[G]} & \ding{51}&\ding{51}&\ding{51}&\ding{55}&\ding{51}&61.3\% &0.9\\
      {[H]} & \ding{51}&\ding{51}&\ding{51}&\ding{51}&\ding{51}&\textbf{65.1\%}&\textbf{0.7}\\
      \bottomrule
    \end{tabular}
    \label{table:ablation}
    \vspace{-2.6em}
  \end{center}
\end{table}

\subsection{Ablation Study}

\textbf{Evaluation of the model  with different designs.}
In Table~\ref{table:ablation}, we conduct extensive ablation studies on the TOSCA dataset to validate the effectiveness of our model designs. 
Specifically, [A] presents a baseline method without any template-related design. 
[B] and [C] demonstrate the employment of the template generation module, assisted respectively through correlation fusion and transitive consistency loss, resulting in accuracy improvements of 2.3\% and 3.5\%. 
The limited improvement is attributed to the difficulty in structural learning of templates without appropriate loss constraints. 
[D] and [E] illustrate the introduction of template construction loss to aid in the learning and optimization of template structures, leading to advancements in accuracy and error compared to settings [B] and [C]. 
Building upon this, [F] and [G] further employ a template selector to identify the most suitable template for aiding the establishment of correspondence for specific shapes. The further improvements in accuracy and error indicate that adaptive template selection significantly contributes to accurate correspondence establishment. 
Finally, in [H], we present the performance of the complete model, highlighting the essential roles played by different modules in collectively achieving accurate point cloud shape correspondences.

\begin{figure}[!htbp]
  \begin{center}
      \includegraphics[width=0.45\textwidth]{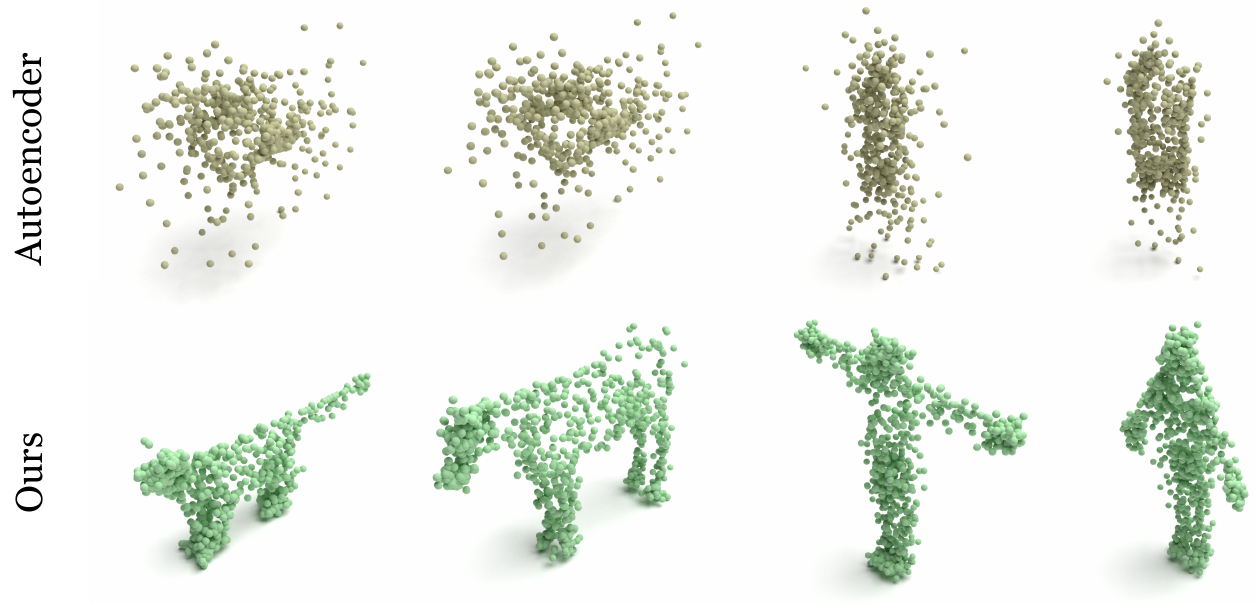}
      \vspace{-0.9em}
      \caption{\textbf{Visual comparison of template generation methods. }  The templates generated by the template generation module exhibit clearer structures in both animal and human templates.}
      \vspace{-1.5em}
      \label{template_vis}
  \end{center}
\end{figure}

\begin{figure*}[!t]
  \vspace{-2.5em}
  \begin{center}
      \includegraphics[width=0.9\textwidth]{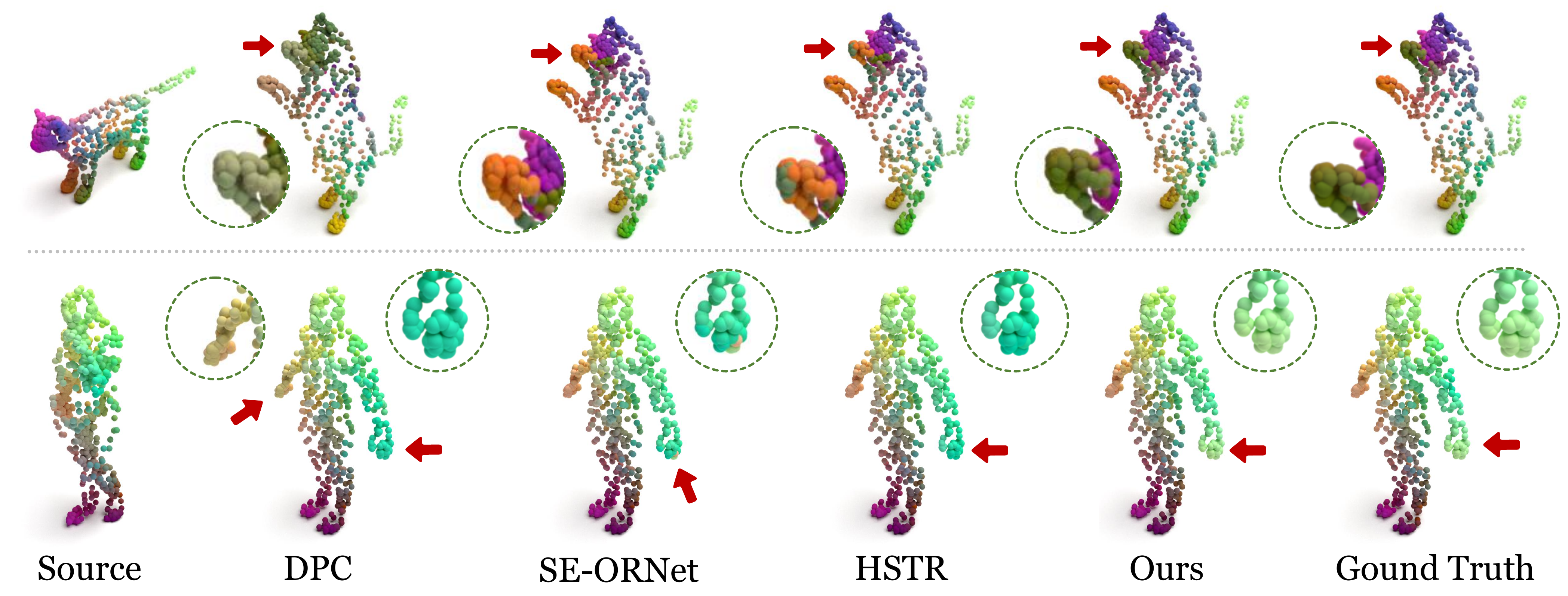}
      \vspace{-1em}
      \caption{\textbf{Visual comparison on TOSCA and SHREC'19. }
      We visually compared the correspondence results of different methods on unconventional shape pairs, providing intuitive validation of the proposed approach's superiority in establishing correspondences.
      }
      \label{vis_compare}
  \end{center}
  \vspace{-1em}
\end{figure*}

\textbf{Effectiveness of the template generation module.}
Figure~\ref{template_vis} compares templates generated by our template generation module and the common autoencoder-based approach~\cite{tian2020shape}. Our designed learnable templates, constrained by the space aligner and construction loss, provide more structural templates to aid point cloud shape correspondence. However, due to the significant shape variations of non-rigid bodies, it remains challenging to generate perfect templates without direct shape supervision. The existing templates are merely a sub-optimal form. Additionally, we carefully consider the number of templates in Table~\ref{table:tem_num}. A single or dual template falls short of capturing the shape variations, resulting in a significant decrease in accuracy. While eight templates can slightly enhance accuracy, they lead to a substantial increase in model parameters. Hence, we offer a trade-off to set up four learnable templates.


\begin{table}[!htbp]
  \begin{minipage}{0.48\textwidth}
    \begin{minipage}[c]{0.48\textwidth}
      \begin{center}
        \footnotesize
        \setlength\tabcolsep{5pt}
        \centering
        \caption{\textbf{Ablation studies on the number of templates. }}
        \vspace{-0.8em}
        \begin{tabular}{c|c|c}
          \toprule
          {\#Template }  & {Param.}  & acc $\uparrow$ \\
          \midrule
          1 & 7.3M &56.8\%\\
          2 & 7.8M &61.4\%\\
          4 & 8.8M &65.1\%\\
          8 & 10.9M &66.2\%\\
          \bottomrule
        \end{tabular}
        \label{table:tem_num}
        \vspace{-1em}
      \end{center}
    \end{minipage}
    \begin{minipage}[c]{0.48\textwidth}
      \begin{center}
        \footnotesize
        \setlength\tabcolsep{5pt}
        \centering
        \caption{\textbf{Comparison of similarity computation methods.} }
        \vspace{0.6em}
        \resizebox{\textwidth}{!}{%
          \begin{tabular}{c|c|cc}
            \toprule
            \multirow{2}*{Settings} &\multirow{2}*{FLOPs(G)} & \multicolumn{2}{c}{TOSCA} \\
            \cline{3-4}
            & & acc $\uparrow$ & err $\downarrow$ \\
            \midrule
            Direct & 1.07 & 65.1\%& 0.73\\
            Transitive & 4.29 &66.3\%& 0.67\\
            \bottomrule
          \end{tabular}
        }
        \label{table:sim_comp}
        \vspace{-0.3em}
      \end{center}
    \end{minipage}
  \end{minipage}
\end{table}

\textbf{Effectiveness of the template assistance module.}
The effectiveness of correlation fusion has been validated in Table~\ref{table:ablation}. Here, we further delve into the discussion regarding the computation of similarities between point cloud pairs. The purpose of transitive consistency loss is to constrain the direct similarity computation between the source and target point clouds to align with the similarity computed through template transitivity. As demonstrated in Table~\ref{table:sim_comp}, similarity computation through template propagation has been proven to be a more stable and accurate approach. However, considering the increased computational complexity associated with the large number of point clouds, we adopt the direct similarity computation method during inference to enhance model efficiency. The transitive consistency loss only constrains the learning of point cloud representations through similarity-guided learning during training.

\begin{table}[!htbp]
  \begin{center}
    \footnotesize
    \setlength\tabcolsep{6pt}
    \caption{\textbf{Cross-dataset generalization evaluation on SMAL and SURREAL benchmarks.} Acc signifies correspondence accuracy at 0.01 error tolerance, and err denotes average correspondence error. The best and second-best outcomes highlighted in bold and underlined, respectively.}
    \label{table:large_dataset}
    \vspace{-0.8em}
    \begin{tabular}{c|c|cc|cc}
      \toprule
      \multirow{2}*{Method} & \multirow{2}*{Input} & \multicolumn{2}{c|}{SMAL} & \multicolumn{2}{c}{SURREAL} \\
      \cline{3-4}\cline{5-6}
      & & acc $\uparrow$ & err $\downarrow$ & acc $\uparrow$ & err $\downarrow$ \\
      \midrule
      {\color{gray}SURFMNet\cite{roufosse2019unsupervised}}   & {\color{gray}Mesh} & {\color{gray}5.9\%} & {\color{gray}0.2}  & {\color{gray}4.3\%} & {\color{gray}0.3} \\
      {\color{gray}GeoFMNet\cite{donati2020deep} }            & {\color{gray}Mesh}  & {\color{gray}/} & {\color{gray}/} & {\color{gray}8.2\%} & {\color{gray}0.2}\\      
      \midrule
      Diff-FMaps\cite{marin2020correspondence}  & Point & / & / & 4.0\% & 7.1 \\
      3D-CODED\cite{groueix20183d}              & Point & 0.5\% & 19.2 & 2.1\% & 8.1 \\
      Elementary\cite{deprelle2019learning}     & Point & 0.5\% & 13.7 & 2.3\% & 7.6 \\
      CorrNet3D\cite{zeng2021corrnet3d}         & Point & 5.3\%  & 9.8 & 6.0\% & 6.9  \\
      DPC\cite{lang2021dpc}                     & Point & 33.2\%  & 5.8 & 17.7\% & 6.1 \\
      SE-ORNet\cite{deng2023se}                 & Point & \underline{36.4\%}  & \underline{3.9} & \textbf{21.5\%} & \textbf{4.6} \\
      HSTR\cite{he2023hierarchical}             & Point & 33.9\%  & 5.6 & 19.4\% & 5.6 \\
      \textbf{Ours}                             & Point & \textbf{37.1\%}  & \textbf{3.7} & \underline{20.6\%} & \underline{4.8} \\
      \bottomrule
    \end{tabular}
    \vspace{-2em}
  \end{center}
\end{table}

\subsection{Cross-dataset Generalization Performance}
The cross-dataset generalization ability is a crucial evaluation metric for the point cloud shape correspondence task. Following the conventional setup~\cite{lang2021dpc,deng2023se,he2023hierarchical}, generalization evaluation involves training on one dataset and testing on another. Specifically, training on the SMAL dataset and testing on the TOSCA dataset, as well as training on the SURREAL dataset and testing on the SHREC'19 dataset. As shown in Table~\ref{table:large_dataset}, our method surpasses existing state-of-the-art approaches, establishing a new state-of-the-art performance by overcoming domain differences on the SMAL dataset. Regarding generalization to the SURREAL dataset, our results are also competitive. Confronting challenges posed by human body symmetry and directional rotations in the SURREAL and SHREC'19 datasets, our approach slightly lags behind the SE-ORNet~\cite{deng2023se}, specifically designed for this issue. However, TANet significantly outperforms other point cloud shape correspondence methods.

\subsection{Visual Comparison }
In Figure~\ref{vis_compare}, we present visual comparisons of our method with competitive approaches. When the source point cloud forms an unconventional shape, such as a distorted standing posture, other methods struggle to distinguish between the hand and facial structures. Our method more accurately establishes the correspondence for hands and heads by leveraging learned templates as an intermediary. 


\begin{figure}[!t]
  \begin{center}
      \includegraphics[width=0.45\textwidth]{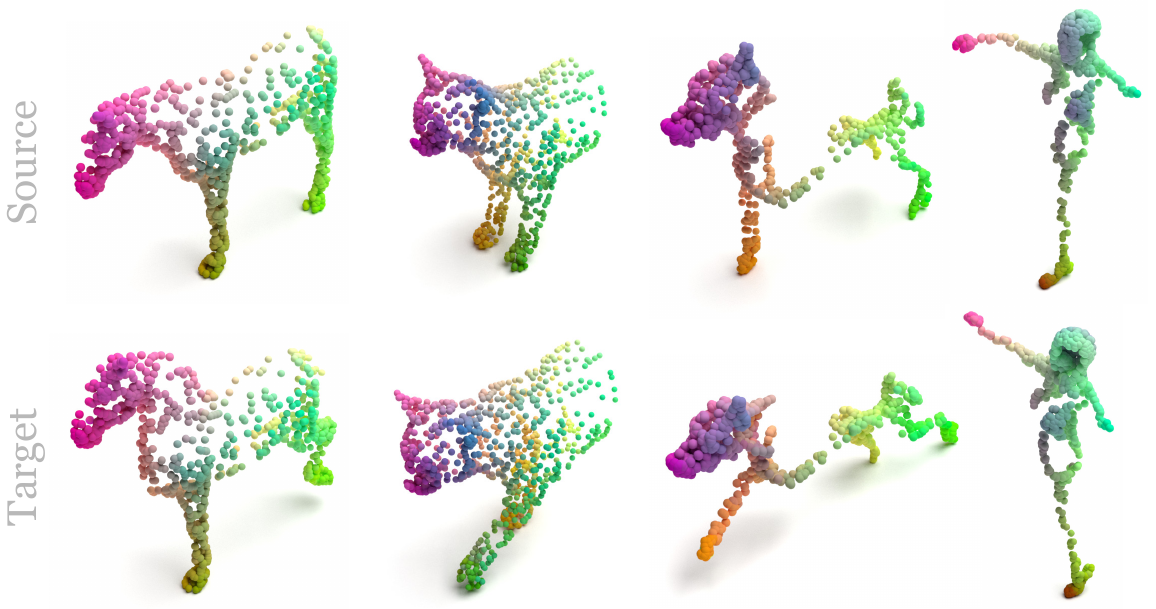}
      \vspace{-0.8em}
      \caption{\textbf{Visualization of the correspondence results on partial-shape SHREC'16 dataset. }
      The partial shapes contain four species, including horses, cats, dogs, and humans. The generation of partial shapes in the SHREC'16 dataset primarily involves two methods: cutting and holing to wipe off body parts. 
      }
      \vspace{-2em}
      \label{shrec16}
  \end{center}
\end{figure}

\subsection{Robustness Analysis}
\label{Robustness_Analysis}
\begin{figure}[!t]
  \begin{center}
      \includegraphics[width=0.30\textwidth]{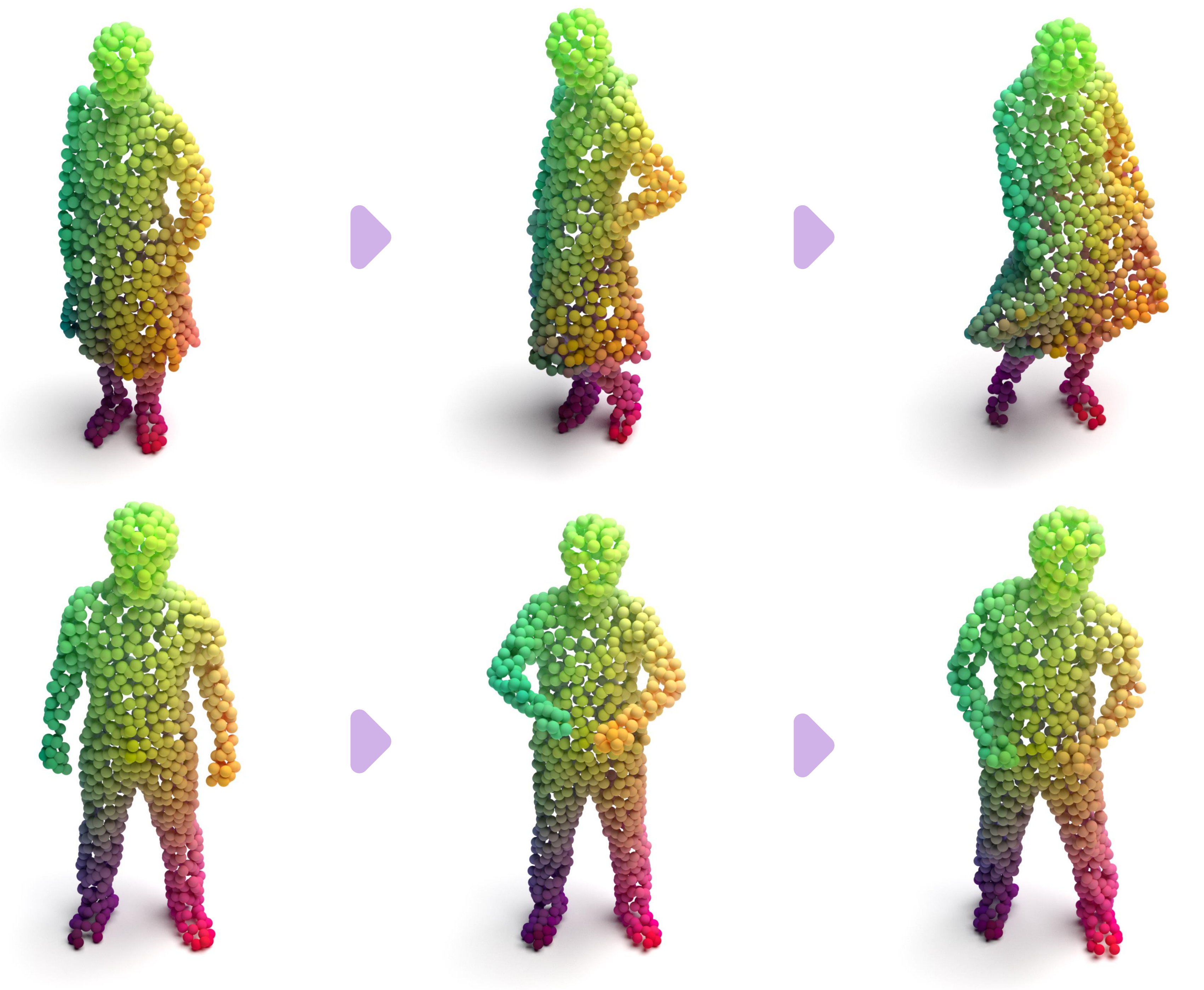}
      \vspace{-1em}
      \caption{\textbf{Visualization of the correspondence results on real scanned Owlii dataset. }
      The results contain two consecutive action sequences involving individuals wearing different attire.
      }
      \vspace{-2.5em}
      \label{owlii}
  \end{center}
\end{figure}

To further validate the robustness and generalization capabilities of our method, we directly evaluate our trained model on the SHREC'16~\cite{cosmo2016shrec} and Owlii~\cite{xu2017owlii} datasets, presenting visual results. As shown in Figure~\ref{shrec16}, the SHREC'16 dataset comprises point cloud pairs of partial shapes, simulating challenges encountered in real-world applications such as occlusion, missing parts, and noise. Although only the upper body of the cat is present, and the dog is represented by a sparse point cloud with minimal parts of its body, our method discriminatively models local structures and accurately establishes shape correspondences. Moreover, the correspondence results of the partial human shape are predictions of the model trained on the animal dataset, providing further validation of the generalization capability.
Illustrated in Figure~\ref{owlii}, the Owlii dataset includes real-scanned human motion sequences, introducing challenges with scan noise and variations in attire and accessories. Nevertheless, TANet accurately establishes the shape correspondences for these sequence point clouds, including details such as dresses worn by the individuals. The visual results on SHREC'16 and Owlii further substantiate TANet's superior robustness and generalization.

\section{Conclusion}
In this paper, we propose an unsupervised template-assisted point cloud shape correspondence network, including a template generation module and a template assistance module. 
Specifically, the template generation module is introduced to learn a set of learnable templates that balance shape complexity and possess an explicitly structured shape.
The template assistance module is designed to leverage the generated templates adaptively to achieve more accurate shape correspondences from template selection, feature learning, and transitive consistency perspectives. 
Extensive experiments on the SHREC'19 and TOSCA benchmarks demonstrate the superiority of TANet. Cross-dataset experiments on SURREAL and SMAL showcase our method's desirable generalization capabilities.

\section{Acknowledgements}
This work was partially supported by the National Defense Basic Scientific Research program (JCKY2022911B002) and the National Nature Science Foundation of China (NSFC 12150007, NSFC 62121002). 

{
    \small
    \bibliographystyle{ieeenat_fullname}
    \bibliography{egbib}
}


\end{document}